\documentclass{article} 
\usepackage{iclr2020templates/iclr2020_conference,times}
\usepackage[english]{babel}

\usepackage{amsmath,amsfonts,bm}

\def\eqref#1{equation~\ref{#1}}

\def\1{\bm{1}}

\DeclareMathAlphabet{\mathsfit}{\encodingdefault}{\sfdefault}{m}{sl}
\SetMathAlphabet{\mathsfit}{bold}{\encodingdefault}{\sfdefault}{bx}{n}

\usepackage{hyperref}
\usepackage{url}
\usepackage{graphicx}
\usepackage{subcaption}
\usepackage{appendix}

\title{Using Simulated Data to Generate Images of Climate Change}

\author{Gautier Cosne*, Adrien Juraver\thanks{equal contribution} ,   M\'elisande Teng*, Victor Schmidt* \\  \textbf{Vahe Vardanyan, Alexandra Luccioni \&  Yoshua Bengio} \\ 
Mila, Universit\'e de Montr\'eal\\
Montr\'eal, Canada}

\iclrfinalcopy 
\begin{document}

\maketitle

\begin{abstract}

Generative adversarial networks (GANs) used in domain adaptation tasks have the ability to generate images that are both realistic and personalized, transforming an input image while maintaining its identifiable characteristics. However, they often require a large quantity of training data to produce high-quality images in a robust way, which limits their usability in cases when access to data is limited. In our paper, we explore the potential of using images from a simulated 3D environment to improve a domain adaptation task carried out by the MUNIT architecture, aiming to use the resulting images to raise awareness of the potential future impacts of climate change.

\end{abstract}

\section{Introduction}
In recent years, many approaches have been proposed for articulating climate change on a personal level, from climate-analog mapping \citep{fitzpatrick2019} to selecting the most impactful and relevant images to communicate the scale of impacts and promising technologies~\citep{chapman2016}. Making the consequences of climate change more concrete can help bridge the ``action-intention gap", i.e. mobilize both individual and collective action. In a related research direction, GANs have been used for humanitarian applications, for instance to increase empathy for victims of far-away disasters by transforming landmarks in major cities as if they were impacted by war~\citep{deepempathy}.

At the nexus of these two approaches, the goal of our project is to raise awareness of how climate change may impact one's personal environment by using GANs to  generate images of these potential impacts. Currently, we are focusing on generating images of one specific extreme climate event: floods. Starting with a street-level image, we use unsupervised image-to-image translation techniques to alter it to reflect the impacts of climate change, projecting flood where it is likely to occur. As is the case for many real-world applications, real data of street-level flooding is scarce and, when available, it lacks many key pieces of information to guide the translation procedure: varied flooded scenes, paired images of the same location before and after flooding, semantic segmentation labels and ground truth scene geometry. To compensate for these data limitations, we have designed a Unity 3D-based simulated world from which we can sample images that have the information lacking in real-world data. Our model, adapted from the MUNIT network~\citep{munit}, is able to leverage both simulated and real images to generate credible flooded scenes.

\section{Our Approach}

\subsection{Real and Simulated Datasets}

The initial dataset we collected for our task consisted of a balanced set of 2000 real images of flooded and non-flooded street-level scenes taken from publicly available datasets such as Mapillary~\citep{mapillary} and Flickr, taken in various weather conditions, seasons, times and viewpoints. While this enabled us to train an initial CycleGAN model~\citep{anonymous1}, we were not able to use it to generate sufficiently realistic images based on this dataset. 
 For this reason, we endeavored to create a simulated world that we could not only use to create more varied training images, but which also gives us the possibility to generate exact labeled pairs of the same location before and after flooding with corresponding geometrical information.

In order to create our simulated world, we used the Unity 3D game engine~\footnote{\url{https://unity.com/}, version 2018.2.21f1}. We created different types of buildings (skyscrapers, individual houses, industrial buildings) in the virtual world, combined with attributes of urban and rural environments: roads, trees, cars, mountains, vegetation, etc.
We developed custom tools to simulate realistic flooding with water reflections and to create paired images of the same exact location before and after a flood. Moreover, for each image captured, we extracted high resolution depth maps, binary masks of floods, and semantic segmentation labels with 10 classes merged from the Cityscapes dataset~\citep{cityscapes}. Our virtual camera parameters (field of view, height etc.) correspond to Google Street View's camera parameters as much as possible, in order to be close to testing conditions. As a starting point, we created 1000 unique pairs of images in the flooded and non-flooded domains for our simulated dataset, with their corresponding depth maps and labels. 
We use both this dataset and the real data in order to train a robust image-to-image translation model, which we describe below.

\subsection{Architecture} \label{sub:archi}
After experimenting with different image-to-image translation networks such as CycleGAN~\citep{anonymous1}, InstaGAN \citep{instagan}, and others and carrying out both quantitative and qualitative evaluations of our results \citep{anonymous1}, we chose to use the MUNIT architecture as the starting point of our model. During our analyses, we found that existing unsupervised GAN evaluation approaches like FID \citep{fid} and KID~\citep{kid} fail to capture the human evaluations of the images we generated, so we continue to rely on empirical evaluations to gauge our model's improvement~\citep{anonymous2}. These evaluations indicate that compared to the other architectures that we tested, MUNIT has a better capacity to generate realistic water textures. 
Furthermore, its approach relies on two discriminators and two generators that attempt to disentangle style and content while performing style transfer by only modifying the style and keeping the content untouched, which is particularly relevant in our context, since flooding does not change the `content' of the image (the street itself, houses, etc). 

\begin{figure}
    \vspace*{-12mm}
    \includegraphics[scale=0.42]{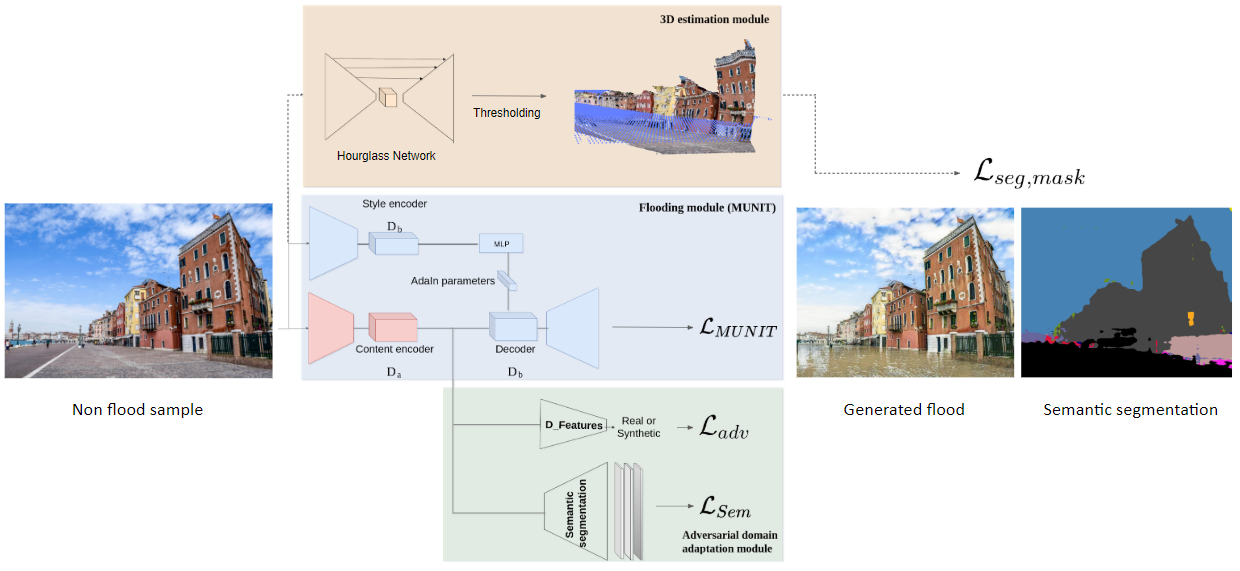}
    \vspace*{-3mm}
    \caption{Model Architecture}
    \label{fig:architect}
    \vspace*{-6mm}
\end{figure}

We made the following changes to MUNIT's architecture for it to be more compatible with our use case (see Figure~\ref{fig:architect}):
\begin{enumerate}
    \item We restricted the network's \emph{Cycle Consistency Loss}~\citep{cyclegan}, which enables the network to reconstruct the entire image following a cyclical A to B (non-flooded to flooded) and B to A (flooded to non-flooded) transformation, so that it is only computed on the part of the image that is not likely to be flooded. This was done to account for the fact that the flooding process constitutes a destructive change, so forcing the network to reconstruct the part of the scene covered by the water would not be relevant for our task. The loss was weakened based on a binary mask of the areas that should be altered in a given image. In the flooded domain, these masks were derived from annotations of the pixels corresponding to water, whereas in the non-flooded domain, these masks were obtained by a 3D reconstruction of the scene (See section \ref{geometry}).
    
    \item We introduced a \emph{Semantic Consistency Loss} to account for the fact that every generated image should keep the same semantic segmentation structure as the source image in all regions of the image except those where modifications should be made (i.e. ground$\rightarrow$water). The semantic segmentation is inferred for both the source and target images using a simplified version of a pre-trained semantic segmentation algorithm, Deeplab v2~\citep{deeplab}. 
\end{enumerate}

\subsection{Leveraging Geometry}
\label{geometry}

Training the MUNIT architecture on real data without consideration of the geometry of the scenes gives us realistic results in terms of the appearance of the water, but does not give us the possibility to determine its location and amplitude. 
Therefore, we introduce binary masks of the areas that should be flooded in a given image. To generate these masks, we recovered the 3D geometry of the street view scenes using the pinhole camera model on images using pseudo relative-depth information obtained using the MegaDepth model \citep{MegaDepthLi18}. Assuming that the water level of the floods corresponds to horizontal planes, we can extract the masks by taking all pixels with height values below a certain fixed threshold -- for instance, the percentage of the image to be flooded. However, MegaDepth does not predict metric depth, such a choice of threshold does not guarantee that the flood level will be consistently plausible on all the images. 
Therefore, we take this approach further and introduce metric information using reference objects (pedestrians, cars, and trucks) for which the real dimensions can be estimated to find the correct scale of the scene. The height estimation of these objects is performed in two steps: detection and mapping. The 3D bounding box detection model of \cite{mousavian2016bounding} is used to estimate the dimensions of the reference objects in the images. Then, the dimensions are matched with the relative coordinates of the objects in the image. Masks are then extracted by specifying a given flood level in the metric system.

\subsection{Learning from Simulated Data}

The goal of our approach is to leverage images from our simulated world to improve the generational capacity of our architecture. However, achieving realistic flooding with a model trained on simulated data does not guarantee that it will maintain its realism when applied to natural images. In fact, in the pixel space there is a domain gap between the distribution of the training set, made of simulated images, and the distribution of the testing set, made of real images. To bridge this gap, we use domain adaptation techniques inspired by research on unsupervised semantic segmentation.

We first investigated an approach involving applying a pixel-level sim2real transformation to all our synthetic images using CyCADA  \citep{cycada}. However, we found that this transformation introduced noise in the obtained pseudo-realistic images: the flooded images generated from this new dataset were less crisp and presented more artifacts than when training on the original simulated data, despite visually coherent pixel-level mapping and the addition of adequate semantic constraints. Thus, we chose to focus on feature space approaches, by implementing an adversarial classifier on the latent space features within the MUNIT architecture. This allows the generator to learn robust high dimensional features that are relevant for the translation task on the source domain and invariant with respect to the shift between the domains. 

Another advantage of using a simulator when the amount of real data is limited is the low cost of noiseless extra information such as ground-truth semantic segmentation and depth information. In our case, we took advantage of these available labels by training a segmentation head that shares the same encoder as the image-to-image generator. We foresee that training this subnetwork will lead to a better structured latent space and facilitate the translation task. We used a modified DeepLab network~\citep{deeplab} as the architecture for the segmentation head and the segmentation labels of our simulated dataset as the training target. Given that we don't have labels in the real world domain, this sub-network is trained with simulated data only.    
\vspace{-1mm}

\section{Results and Discussion}
\begin{figure}
    \vspace*{-10mm}
    \includegraphics[scale=0.242]{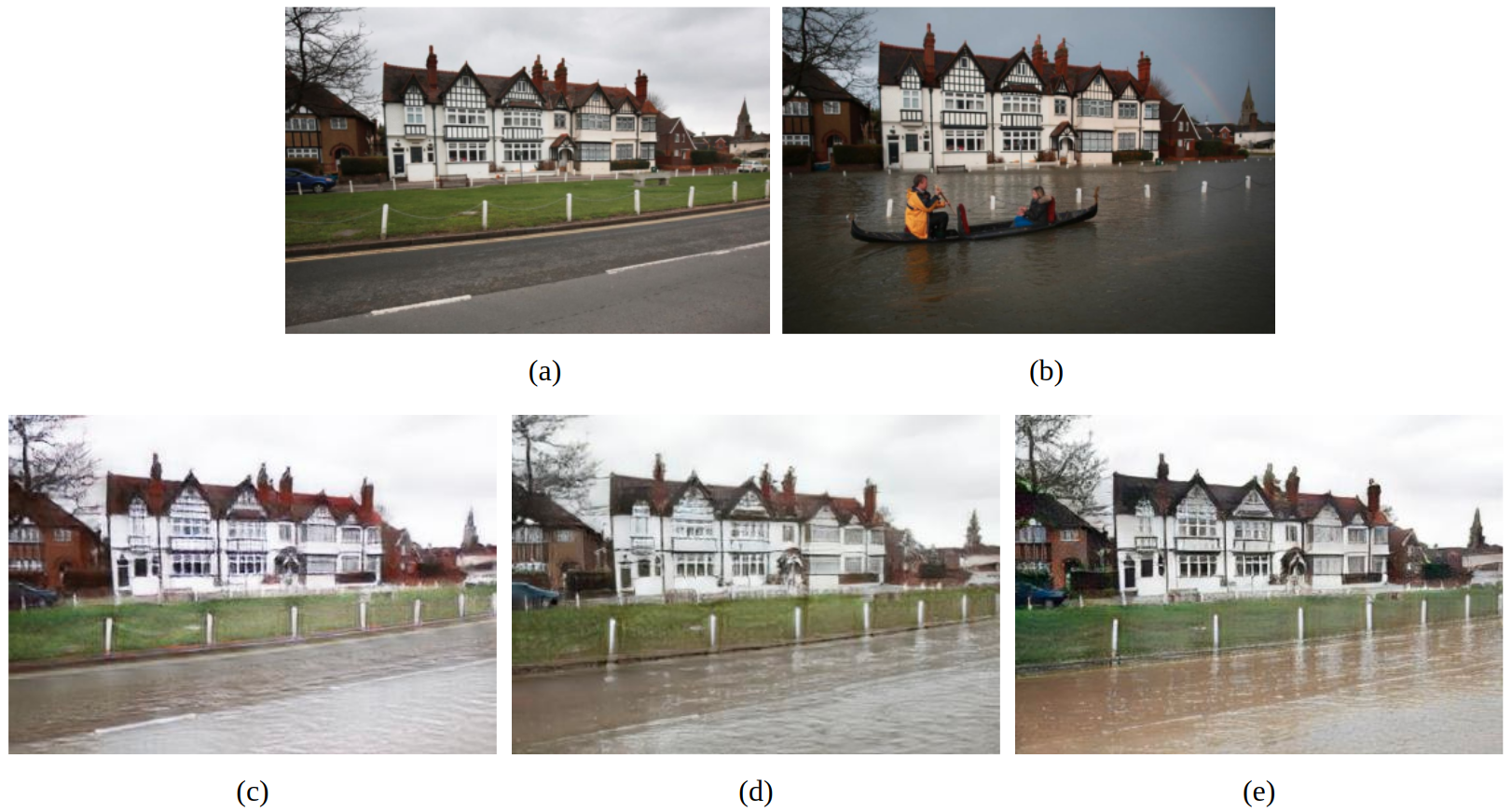}
    \vspace*{-1mm}
    \caption{Illustrations of our results with different methods. (a) Is a natural image.  (b) Is the same scene captured after a flood. (c),(d) and (e) are three generated images with a model trained on: real data (c), simulated and real data without domain adaptation (d) and with domain adaptation (e). 
    }
    \vspace*{-3mm}
    \label{fig:results}
\end{figure}

In the previous section, we presented an approach that leverages both real and simulated data to perform image-to-image translation to simulate the impacts of climate change in the real world. With domain adaptation techniques, we are able to efficiently use two different sources of unaligned data to achieve a quality of results that is superior to one that does not leverage simulated data. The addition of an adversarial learning classifier and a segmentation head enables us to take advantage of the unlimited amount of synthetic pairs and labels offered by the simulator to generate more realistic images of floods in street-level scenes. In Figure~\ref{fig:results} (c), we show the results of a model trained only on real data, compared to (d) and (e), which leverage both real and simulated data -- it can be observed that there is a blur on the house and trees, and the rendered water is less opaque and reflective, resembling more cement than actual floodwater and retaining the characteristics of the original road (markings, etc). Furthermore, the changes that we made to the MUNIT architecture (see Subsection~\ref{sub:archi}) enhanced the generated image further, improving both the position of the water in the image, limiting the flooding to plausible areas without degrading the rest of the image, as well as improving the quality and realism of the water generated. This can be seen comparing the images in Figure~\ref{fig:results} (d) and (e) : without domain adaptation, the model trained on simulated and real data (d) performs worse than when trained on real data only (c), whereas with domain adaptation (e), the details of the house and trees are preserved, and the water looks more realistic.  

We are continuing work on our project to generate images of flood impacts  to raise awareness of the future impacts of climate change. In terms of improving the performance of our current model, we aim to be able to condition the generation based on a given flood level; to do so, we plan on leveraging geometric information from the simulator and training an end-to-end height estimator in order to compute binary masks of the areas to flood that are consistent with the geometry of the scene, even when camera parameters are not available. The domain adaptation module can also be improved by adding an adversarial classifier in the output space of the segmentation head as in~\citep{outputda}. We also plan to represent other weather events such as wildfires and droughts while continuing to leverage both real and simulated data. Indeed, our simulated dataset can be easily augmented with other effects such as fire while producing the same capture points and metadata. We aim to develop an interactive website that, given a user-entered address, will query the Google Street View API~\citep{streetview} to get an image of the location and alter it to display a plausible image of its climate future based on the predictions of climate models. We hope this tool will help communicate effectively on climate change related risks.

\newpage
\section{Acknowledgements}
We thank Sahil Bansal for his contribution on the creation and annotation of our real images dataset and initial tests to select the model architecture. We thank L\'eopold Herlaud for his work on lighting and compositing and the creation of shaders for the simulated dataset. We are grateful to Felipe Codevilla for advice on the sensing modalities of our simulator. 
We would also like to give thanks to Microsoft Research for helpful discussions and feedback.

\bibliography{ccai}
\bibliographystyle{iclr2020templates/iclr2020_conference}

\newpage
\appendix
\appendixpage
\addappheadtotoc
\section{End-To-End Height Estimation}
We propose to leverage the known geometry of simulated data and to build a pixel-wise height predictor by extracting depths maps directly from our simulator and computing ground truth height maps using the pinhole camera model. 
Indeed, training images and flood masks for the flood location constraint on the non-flooded to flooded domain translation can be generated in the simulator for any chosen height, something which is not directly available on our dataset of real-world images acquired in the wild.While methods for single image depth estimation have been investigated for many years~\citep{depth_est,cnn_depth_estimation, MegaDepthLi18}, we have found no work in height estimation from street view scenes images. Following the success of \cite{MegaDepthLi18} on the task of depth estimation, we train an hourglass network \citep{hourglass} to predict metric height.
In a first step, we use an L2 loss function with a mask on the pixels corresponding to the sky. 
The first experiments, which involved training separate models on a dataset of 8500 images extracted from the CARLA dataset~\citep{Dosovitskiy17} (6500 training, 2000 testing) and the non-flooded domain dataset from our simulator produced promising results. The output height maps still need to be smoothed, as the edges determining changes in texture/color result in a higher gradient of height in our predictions than should be the case.  We are also investigating how such models can transfer to real images. 

\section{Failure and Success cases}
\begin{figure}[!h]
    \includegraphics[scale=0.28]{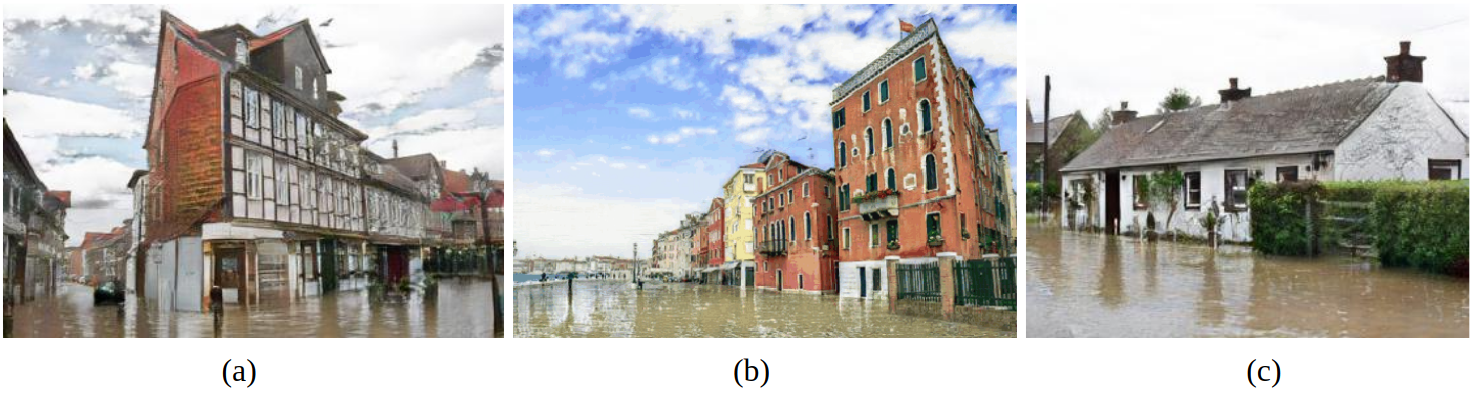}
    \vspace*{-3mm}
    \caption{Success cases after flooding. (a) Image with multiple streets, (b) Image in urban area  (c) Image with single house.}
    \label{fig:success}
\end{figure}

There are several limitations and cases where our current model does not work as expected. Often when there is grass on the side of the road, it is not adequately flooded: the color doesn't change, or the algorithm circumvents the zone in its alteration process.
Another failure case happens when we infer our model on street-view images for which the camera viewpoint is closer than usual to the building. In the latter case, we observed that the images, presenting less road area, are more prone to be less realistically flooded as if the algorithm has lost its bearings regarding the region that should be submerged. On the contrary, images where a large part is covered by a uniform ground (uni-color or with a unique ground texture) are well flooded (Figure~\ref{fig:success}). 

\begin{figure}[!h]
    \includegraphics[scale=0.32]{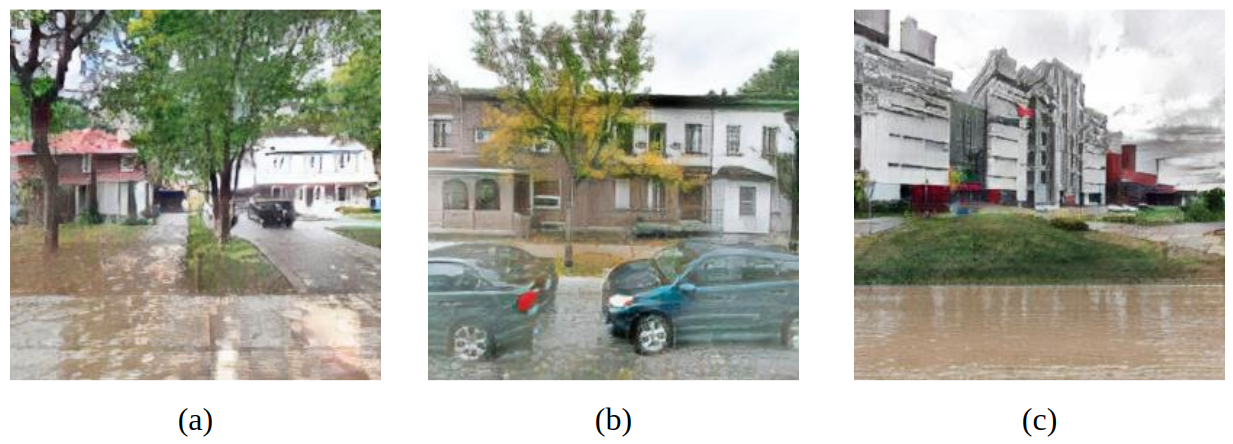}
    \vspace*{-3mm}
    \caption{Failure cases after flooding. (a) Image without road, (b) Image with a small road region and cars  (c) Image with grass.}
    \label{fig:failures}
\end{figure}

\pagebreak

\section{Description of the virtual world}

\begin{figure}[h!]
    \begin{centering}

    \includegraphics[width=\textwidth]{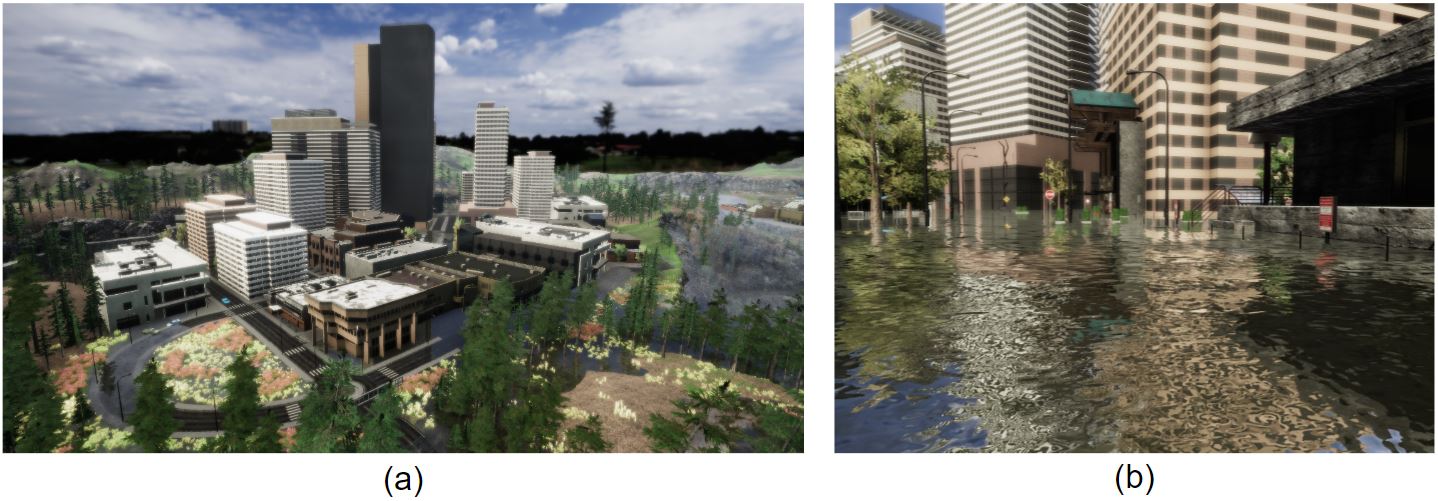}
    \vspace*{-3mm}
    \caption{(a) Bird's eye view of the virtual city, (b) Flooded scene using reflective water shader}
    \label{fig:mergeUnity}
    \end{centering}
    \vspace*{-6mm}
\end{figure}

\begin{figure}
    \includegraphics[width=1.0\textwidth]{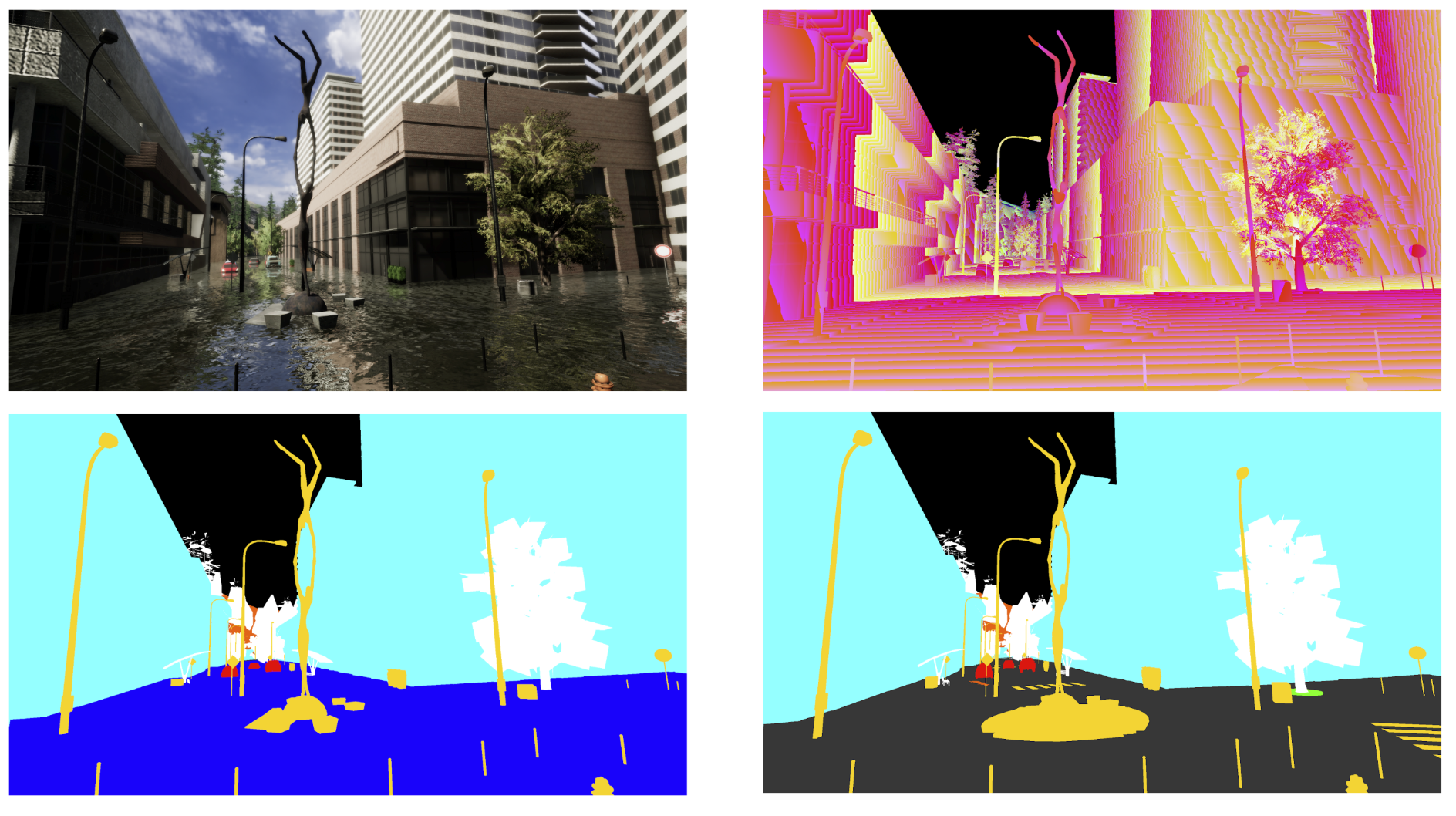}
    \vspace*{-3mm}
    \caption{Depth map and semantic segmentation produced in the simulated world}
    \label{fig:labels}

\end{figure}

The prototype and starting point of the virtual world is taken from Microsoft's AirSim project, designed for drone flight simulations~\footnote{\url{https://github.com/microsoft/AirSim}}. We have modified and adopted the scene and assets to correspond to our needs. We have also made it almost four times bigger than its initial size -- the overall size of our virtual world is now about 1km$^2$ (Figure~\ref{fig:mergeUnity})(a). The custom reflective water shader helps to render the simulated water to be as close to real world flooded scenarios as possible (Figure~\ref{fig:mergeUnity})(b).
Finally, we have modified Unity's built-in tools to capture identical paired images of the same view with and without flood. All animations and engine's main update loop were stopped to be able to capture a sufficient level of pixel-wise similarity between flooded and non-flooded versions of the same spot. In order to render the binary mask, depth map and semantic segmentation image, we added additional virtual cameras that have exact same parameters as the main camera to match the original captured image. 
Binary masks are generated with a simple water/no-water pixel checking operation, where water becomes a white pixel and everything else is black. To save the depth map, we used the GPU depth buffer and transferred the values into 3-channel RGB image with high resolution that cover up to 655.36m (Figure~\ref{fig:labels}: top-right). We have created custom replacement shader to capture ground-truth semantic segmentation images both for flooded and non-flooded versions (Figure~\ref{fig:labels}: bottom left and right). We also save all necessary metadata in a JSON file to be able to revisit the same exact data points in the future, for example with different weather conditions, different luminosity parameters of scene or to simply augment the dataset if needed. The resolution of images are 2560x1440, which is higher than the average resolution of real world images that we gathered.

\end{document}